# Learning to attend in a brain-inspired deep neural network

**Hossein Adeli (hossein.adelijelodar@gmail.com)**
Department of Psychology
Stony Brook University, NY

**Gregory Zelinsky (gregory.zelinsky@stonybrook.edu)**
Departments of Psychology and Computer Science
Stony Brook University, NY

**Abstract:**

**Recent machine learning models have shown that including attention as a component results in improved model accuracy and interpretability, despite the concept of attention in these approaches only loosely approximating the brain's attention mechanism. Here we extend this work by building a more brain-inspired deep network model of the primate ATTention Network (ATTNet) that** *learns to shift its attention so as to maximize the reward.* **Using deep reinforcement learning, ATTNet learned to shift its attention to the visual features of a target category in the context of a search task. ATTNet's dorsal layers also learned to prioritize these shifts of attention so as to maximize success of the ventral pathway classification and receive greater reward. Model behavior was tested against the fixations made by subjects searching images for the same cued category. Both subjects and ATTNet showed evidence for attention being preferentially directed to target goals, behaviorally measured as oculomotor guidance to targets. More fundamentally, ATTNet learned to shift its attention to target like objects and spatially route its visual inputs to accomplish the task. This work makes a step toward a better understanding of the role of attention in the brain and other computational systems.**

## Introduction

Visual attention enables primates to prioritize the selection and further processing of visual inputs for the purpose of achieving behavioral goals, but how is this attention control learned? Most neural and cognitive models avoid asking this question, focusing instead on the *effects* that prioritization and selection have on neural and behavioral responses. For example, neural models of attention control are largely based on Biased Competition Theory (Desimone & Duncan, 1995), which focuses on the effects that attending to an object's location or features has on neural recordings and brain network dynamics (Deco & Zihl, 2001; Hamker, 2004; Reynolds & Heeger, 2009). Similarly, cognitive computational models of attention aim to predict the selection and guidance of attention shifts to behavioral goals using image-computable methods and visually-complex inputs (Adeli, Vitu, & Zelinsky, 2017; Itti & Borji, 2015; Itti & Koch, 2000; Miconi, Groomes, & Kreiman, 2015; Tsotsos et al., 1995; Zelinsky, Adeli, Peng, & Samaras, 2013). These neural and cognitive models of attention control are therefore engineered to fit (or predict) behavior and/or neural data without addressing the more fundamental questions of how attention control signals emerge and function in the context of performing a task (Gottlieb, Hayhoe, Hikosaka, & Rangel, 2014; Gottlieb, Oudeyer, Lopes, & Baranes, 2013; Hayhoe & Ballard, 2014; Tatler, Hayhoe, Land, & Ballard, 2011) or why the brain might even find prioritizing visual inputs and shifting attention to be a useful thing to do. Recent methods in machine learning have engaged these difficult questions, showing that models able to learn to shift an attention focus yield improved accuracy and interpretability in applications ranging from object classification (Wei, Adeli, Zelinsky, Hoai, & Samaras, 2016) and detection (Mnih, Heess, & Graves, 2014) to caption generation (Xu et al., 2015) and language translation (Vaswani et al., 2017). However, the focus of these models was on performance, and not on testing against human attention behavior. Their designs were also not informed, beyond a broad concept of "attending", by cognitive and neural findings on primate attention mechanisms. Here we leverage the potential of these three perspectives by introducing ATTNet, an image-computable DNN model of the ATTention Network. ATTNet is inspired by biased-competition theory and trained using deep reinforcement learning. Through the application of reward in the context of a search task (Fig. 1a), ATTNet learns to shift its attention to the locations of features of the rewarded object category.

## Methods

ATTNet consists of three interacting components: (1) early parallel visual processing (2) ventral processing, and (3) dorsal processing. This organization in based on an influential characterization of visual processing in the cortex on the division of labor between the two visual pathways diverging after the initial parallel processing of visual input (Ungerleider & Haxby, 1994; Ungerleider & Pessoa, 2008).

In ATTNet, the initial stage is modeled by the convolutional layers of a Convolutional Neural Network (CNN) trained for object classification (Fig 1b, right)). CNN architecture is itself brain-inspired, particularly the hierarchical architecture of the mammalian visual processing (Fukushima, 1980; LeCun et al., 1989). Given the success of CNNs in pattern recognition, their brain-inspired hierarchical architecture, and recent work showing the similarity between representations built across a CNN's layers to those of brain areas in the ventral pathway (Cadieu et al., 2014; Cichy, Khosla, Pantazis, Torralba, & Oliva, 2016; Güçlü & van Gerven, 2015; Khaligh-Razavi & Kriegeskorte, 2014; Yamins et al., 2014), they are well suited as models of representation learning in the visual system. The input image will be fed into a commonly used image classification CNN, VGG16 (Simonyan & Zisserman, 2014). The convolutional layers of this network grossly approximate visual processing in areas V1 to V4, with the output of the final convolutional layer being the V4 activation. As shown in figure 1b this output is a 14*14*512 pixel activation map, which codes the filter responses from each of 512 relatively high-level features over a coarse but spatially-organized 14*14 pixel map of visual space. This is assumed to be the rich bottom-up representation of a visual input that is produced by parallel early visual processing.

The ventral "what" pathway, long assumed to endow primates with their ability to recognize objects and scenes (Felleman & Van Essen, 1991; Ungerleider & Haxby, 1994), extends temporally, from early visual areas, to Posterior inferotemporal (PIT) and then Anterior inferotemporal (AIT) cortex (IT). Under ATTNet, positioning of V4 between early visual and ventral (and dorsal) processing makes it a key attention control structure, which is in line with the literature showing that effects of attention are most strongly observed in this area (Bichot, Rossi, & Desimone, 2005; Chelazzi, Miller, Duncan, & Desimone, 2001). Our premise is that the direction of attention to a location in V4's activation map determines how information from visual inputs is routed to the IT structures for the purpose of improving object classification success. This *selective routing* is modeled as a 4*4 selection window centered at the attended location (Fig. 1d; red rectangle). ATTNet models IT processing using two trainable fully-connected layers, with the output feeding into a prefrontal layer that eventually makes the decision on whether the target object was present or absent.

The dorsal "where" pathway extends dorsally from early visual processing into Posterior Parietal Cortex (PPC), long believed to be responsible for the spatial prioritization of visual inputs and the guidance of actions to objects (Bisley & Goldberg, 2010; Ptak, 2012; Szczepanski, Pinsk, Douglas, Kastner, & Saalmann, 2013). ATTNet's dorsal network serves the same function; it spatially prioritizes and selects visual inputs for selective routing through the ventral pathway, thereby imposing seriality on behavior requiring confident classification decisions. Dorsal processing is modeled using two PPC layers (Fig. 1b). PPC1 takes input from a wide-angle window of the V4 activation map (Fig. 1d, green box), which has coarse retinotopy, and learns to weight the input to create a priority map indicating evidence for the target category in the scene. PPC2 then combines this priority map with another map holding the locations of all previously attended areas, and from this combined map the most active location is selected and used to position the next ventral selective routing window. This cycle of prioritization and selection repeats 5 times, with PFC summing the 5 ventral outputs to make the target present/absent decision.

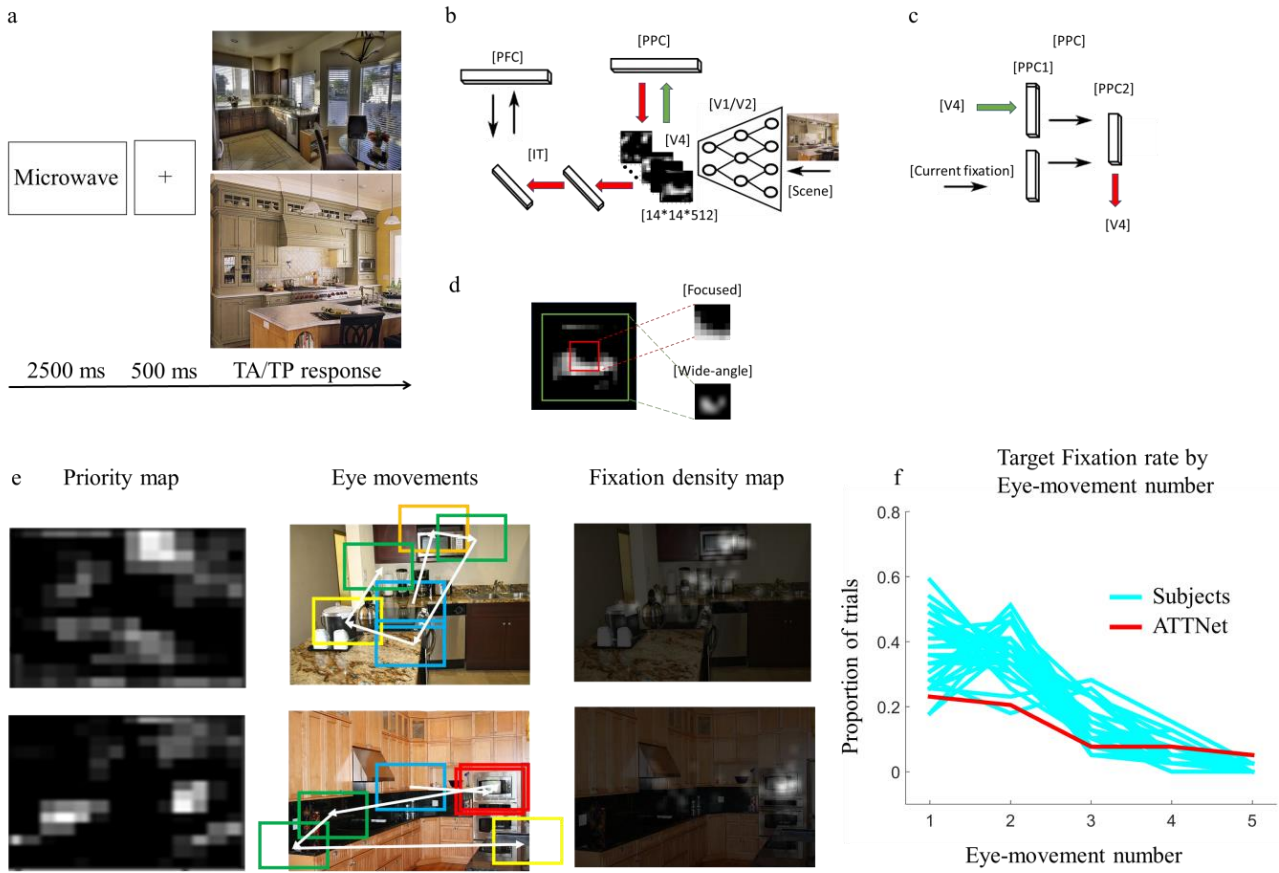

Figure 1. **(a)** Behavioral experiment procedure. **(b)** Anatomy of ATTNet. **(c)** ATTNet's PPC. **(d)** Routing windows. **(e)** ATTNet's eye-movements on two sample search displays with the corresponding dorsal priority and fixation density maps. **(f)** Guidance to the target quantified as the proportion of trials where target was fixated at each eye-movement.

ATTNet was trained using policy gradient reinforcement learning (Mnih et al., 2014; Williams, 1992) to respond target present or target absent for a microwave target (category selected to have minimal center bias in the test set) in 2000 kitchen scenes (half target present) from COCO (Lin et al., 2014). A behavioral ground truth was obtained by having 30 subjects perform present/absent categorical search for a microwave target (Fig. 1a) in 80 images, also from COCO but a disjoint set from those used during training.

## Results

Figure 1e shows the process of prioritization-selection-routing-classification. Like subjects, ATTNet started each trial fixating at the center of the scene. The information from this area is routed along the ventral and dorsal pathways (the boxes show the visual areas that are routed ventrally). The priority map generated in the dorsal pathway guides attention to a new location and the process repeats. The model is only trained on the overall target present and target absent judgment without prior knowledge on the target object category. But over the course of the training the model detects that certain patterns are rewarding and attends to these patterns to be able to make a more informed decision. Left column shows the priority map that is generated at the initial fixation in the dorsal pathway for the two sample displays indicating that the model learns to bias the visual space for visual input reflecting target-category. The routing windows are colored based on the ventral response from blue to green to red with warmer colors showing more confidence in the routed pattern being the target. The right column on figure 1e shows the fixation density maps from the 30 subjects for comparison to the priority map.

To quantify ATTNet's attention being preferentially guided to the target, we plot the proportion of trials where the target was first fixated at each eye-movement for ATTNet (red) and individual subjects (cyan) in figure 1f. As shown in this plot, while not as strong as the subjects, ATTNet shows guidance to the target by preferentially fixating the target in earlier eye-movements.

## Discussion

Understanding computational principles of selective attention is key to understanding brain function and building brain-inspired AI systems (Hassabis, Kumaran, Summerfield, & Botvinick, 2017; Lake, Ullman, Tenenbaum, & Gershman, 2017; Marblestone, Wayne, & Kording, 2016). This work studies the computational benefit of attention as a dynamic selective routing of information for performing a difficult visual search task.